\documentclass{article}
\usepackage[T1]{fontenc}
\usepackage{iclr2027_conference,times}
\usepackage{amsmath,amssymb,amsfonts}
\usepackage{graphicx}
\usepackage{booktabs}
\usepackage{microtype}
\usepackage{xcolor}
\usepackage{url}

\title{\textsc{SPT-Flow}: One Integration, Many Futures\\in Continuous-Time World Models}

\author{\hspace*{-\tabcolsep}\parbox[t]{\textwidth}{\centering
\setlength{\parskip}{0pt}
{\normalfont\fontsize{11}{16}\selectfont\bfseries
Mingyang Lyu$^{1,2,5,*}$\hspace{1.2em}Yinqian Sun$^{1,*}$\hspace{1.2em}Gongce Cao$^{2,*}$ \\[1pt]
Yuhao Zhang$^{3}$\hspace{1.2em}Lu Jia$^{4}$\hspace{1.2em}Qian Zhang$^{1}$ \\[1pt]
Bo Wang$^{1,\dagger}$\hspace{1.2em}Pin Lyu$^{1,\dagger}$\endgraf}
\vspace{8pt}
{\normalfont\fontsize{9}{12}\selectfont $^{1}$ Institute of Automation, Chinese Academy of Sciences \\
$^{2}$ University of the Chinese Academy of Sciences \\
$^{3}$ Xi'an Jiaotong University\qquad $^{4}$ University of Science and Technology Beijing \\
$^{5}$ Zhongguancun Academy\endgraf}
\vspace{6pt}
{\normalfont\fontsize{8.5}{10}\selectfont $^*$ Equal contribution.\qquad $^{\dagger}$ Corresponding authors.\endgraf}
}\hspace*{-\tabcolsep}}
\iclrfinalcopy

\newcommand{\method}{\textsc{SPT-Flow}}

\begin{document}
\maketitle
\suppressfloats[t]

\begin{abstract}
Continuous-time world models generate a latent trajectory through a single
ODE integration, but adding stochastic conditioning does not ensure that the
model uses it. In our teacher-forced baseline, posterior means vary by only
about $1\%$ of the prior scale, and matched and swapped latents yield
indistinguishable reconstruction PSNR at the reported precision. We study a
supervision-shortcut hypothesis: ground-truth intermediate states can reduce
the need for trajectory-associated latent information. We introduce
\method{}, which samples one latent per trajectory and matches a free-running
ODE rollout to demonstration states at multiple readout times. Re-anchoring
experiments support the proposed mechanism. On LIBERO, with both models
trained for $30$k optimizer steps, \method{} improves best-of-10 PSNR over our
retrained ODEWorld baseline by $2.0$ and $2.4$\,dB at the mid-trajectory and
final frames and reduces mean LPIPS by $23\%$, with comparable per-sample
inference cost. The rollout control recovers most of the accuracy gain;
learned latent conditioning adds diversity and best-of-$K$ gains while
reducing mean sample accuracy relative to this control. The diagnostics
support trajectory-associated latent utilization; semantic mode coverage
under matched conditions in the robotic setting remains unverified.
\end{abstract}

\section{Introduction}

World models support robotic manipulation by predicting future observations
conditioned on an initial observation and a goal
\citep{du2023unipi,black2024susie,hu2025vpp}. Many approaches use discrete
stochastic \citep{ha2018world,hafner2019planet,hafner2025dreamerv3} or
deterministic \citep{hansen2024tdmpc2,zhou2025dinowm} latent transitions.
ODEWorld \citep{liu2026odeworld} instead models latent dynamics as an ordinary
differential equation in physical time, generating a trajectory through a
single numerical integration with arbitrary-time readout.

\begin{figure}[t]
\centering
\includegraphics[width=\linewidth]{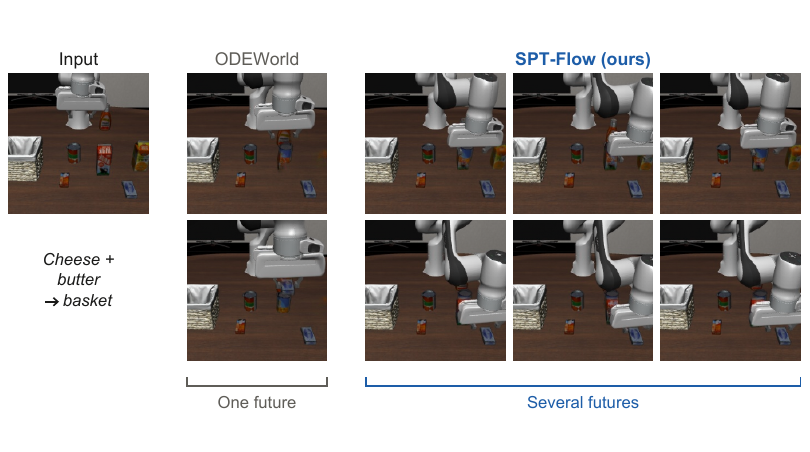}
\caption{Predictions conditioned on the same initial frame and instruction.
ODEWorld produces one trajectory; \method{} samples three, each through one
integration. Rows show $t/T=0.6$ and $1.0$. The example is selected for
diversity; model settings follow Figure~\ref{fig:qual}.}
\label{fig:teaser}
\end{figure}

This formulation nevertheless produces only one trajectory for a fixed initial
state and goal. Manipulation tasks can admit multiple valid continuations:
\emph{put both pots on the stove}, for example, permits either pot to be moved
first. When the velocity-field inputs cannot distinguish these alternatives,
squared-error supervision favors their conditional mean, whose integration may
match neither continuation. Appendix~\ref{app:toy} illustrates this failure in
a controlled setting and presents distinct continuations recorded on a physical
robot arm.

Conditional latent variables offer a standard approach to multimodal
prediction \citep{sohn2015cvae}, but improved accuracy does not guarantee
informative latent representations. In our experiments, adding a
trajectory-level latent $\epsilon$ improves best-of-$K$ accuracy while
posterior means vary little across trajectories and swapping inferred latents
barely affects reconstruction accuracy (Section~\ref{sec:diag}). We term this
behavior \emph{decorative latent conditioning}.

A potential explanation is a supervision shortcut: ground-truth intermediate
states can reveal the demonstrated continuation, allowing the velocity field
to predict its target without using $\epsilon$. At inference, these states are
unavailable, yet training has not explicitly required the latent to resolve
the remaining ambiguity.

We address this mismatch with \method{}, which replaces teacher-forced
velocity supervision with rollout matching. During training, a latent sampled
from the approximate posterior conditions a free rollout from $z_0$, whose
predicted states are matched to the demonstration. Removing intermediate
ground-truth inputs encourages the model to encode and use
trajectory-associated information through $\epsilon$. At inference, different
prior samples generate different trajectories, each through a single
integration (Figure~\ref{fig:teaser}). Here, multiple futures denotes sampled
trajectories; the LIBERO diagnostics assess latent utilization, while recovery
of known continuation modes is tested in the synthetic setting of
Appendix~\ref{app:toy}.

\paragraph{Contributions.}
\label{sec:intro-contrib}
\begin{itemize}
\item We introduce \method{}, combining trajectory-level latent conditioning
with rollout matching for stochastic physical-time prediction. It preserves the
deterministic baseline's number of velocity-field evaluations per trajectory
and achieves comparable inference latency.
\item We characterize decorative latent conditioning using posterior
specificity and latent swapping. Under rollout matching, specificity exceeds
the predefined threshold in all three seeds, and matched latents outperform
swapped latents by $1.05$\,dB. Controlled re-anchoring experiments provide
evidence for the proposed supervision shortcut.
\item On LIBERO, at $30$k optimizer steps, \method{} improves best-of-10 PSNR over our retrained
ODEWorld baseline by $2.0$ and $2.4$\,dB, and mean sample PSNR by $1.0$ and
$1.3$\,dB, at the mid-trajectory and final frames, respectively, while
reducing mean LPIPS by $23\%$. The rollout control recovers most of the
accuracy gain; learned latent conditioning improves best-of-$K$ performance
and diversity at the expense of mean sample accuracy relative to that control. The
prediction advantage persists when both methods receive four times the
standard training steps.
\end{itemize}

\begin{figure}[t]
\centering
\includegraphics[width=\linewidth]{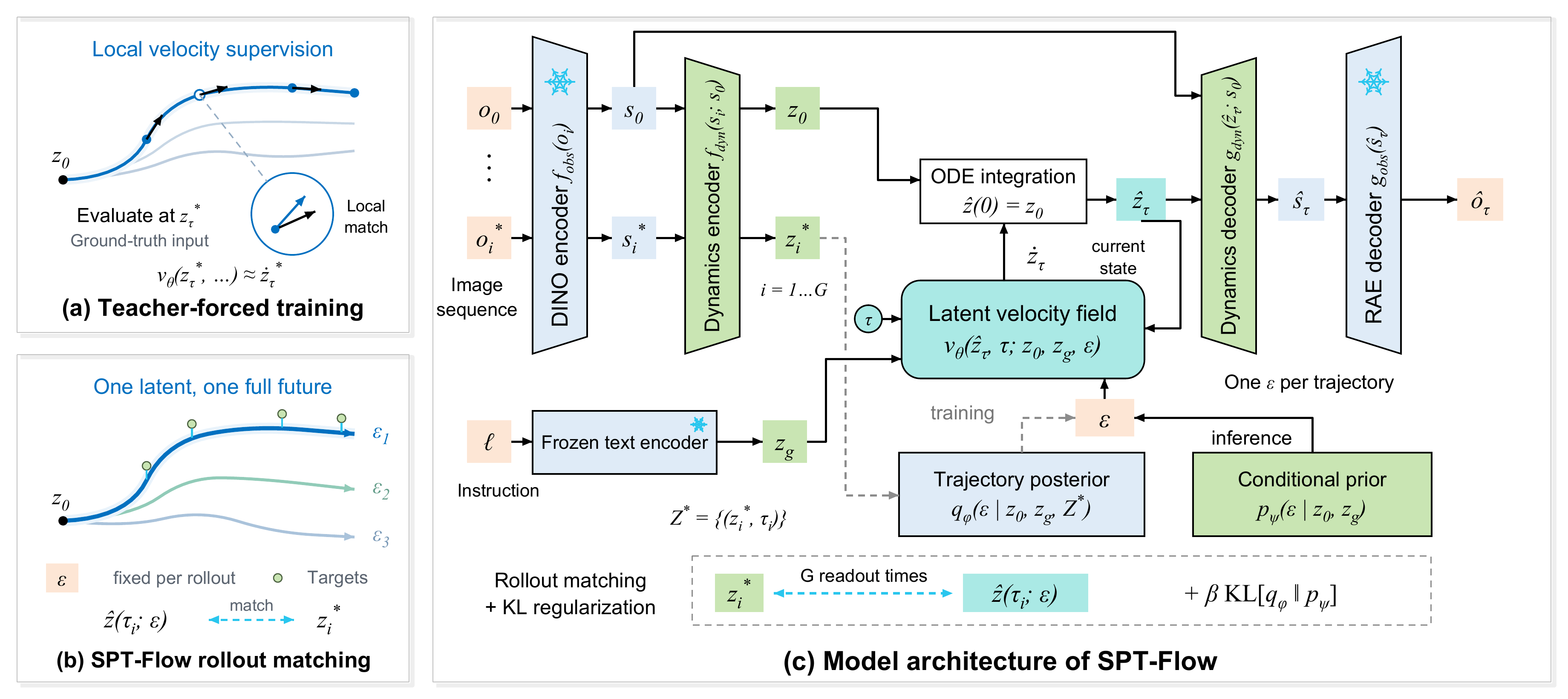}
\caption{\method{} and its training objective. \textbf{(a)} Teacher-forced
velocity supervision (Equation~\ref{eq:tf}) evaluates the field at a
ground-truth latent $z^{\star}_{\tau}$, which can reveal the demonstrated
continuation. \textbf{(b)} Rollout matching (Equation~\ref{eq:roll}) samples
one $\epsilon$, integrates from $z_0$, and supervises predictions at $G$
readout times without intermediate ground-truth inputs. \textbf{(c)}
Architecture: snowflakes indicate frozen modules and dashed paths are used
only during training. At inference, $\epsilon$ is sampled from $p_\psi(\epsilon
\mid z_0, z_g)$.}
\label{fig:method}
\end{figure}

\section{Related work}
\label{sec:related}

\paragraph{World models.}
World models support manipulation through goal-conditioned visual prediction
\citep{du2023unipi,black2024susie,hu2025vpp,huang2025ladiwm}. Many approaches
use discrete stochastic \citep{ha2018world,hafner2019planet,hafner2025dreamerv3}
or deterministic \citep{hansen2024tdmpc2,zhou2025dinowm} latent transitions.
Continuous-time alternatives include neural ODEs \citep{chen2018neuralode},
latent ODEs \citep{rubanova2019latentode}, neural SDEs
\citep{li2020scalablesde}, and continuous-time video models
\citep{park2021vidode,ye2024stdiff}. ODEWorld \citep{liu2026odeworld} generates
latent trajectories by integration in physical time.\footnote{We distinguish
posterior collapse of a variational latent from the representation collapse
discussed in ODEWorld.} Recent stochastic world models include object-centric
latent dynamics \citep{daniel2026lpwm} and flow matching in pretrained feature
spaces \citep{porcher2026flowwm}. \method{} builds on ODEWorld's representation
and single-integration formulation, introducing trajectory-level latent
conditioning trained through rollout matching.

\paragraph{Latent conditioning and utilization.}
Under standard uniqueness conditions, deterministic ODE dynamics produce one
trajectory for a fixed initial state and conditioning
\citep{dupont2019anode}. Latent conditioning provides one way to represent
multiple futures, as explored in neural ODEs \citep{norcliffe2021ndp,cheng2024latentebmode} and
stochastic video prediction
\citep{babaeizadeh2018sv2p,lee2018savp,denton2018svg,franceschi2020srvp,ye2019cvp}.
The same argument underlies generative action heads, where a sample from a
diffusion or flow-matching policy selects an action chunk instead of
regressing to the mean of the demonstrated ones
\citep{chi2023diffusionpolicy,black2025pi0,hu2024adaflow,zhai2025vfp}, and
\citet{guo2025vrfm} attach a latent to a flow-matching field for the same
reason in generative flow time.
However, variational latents may remain uninformative because of optimization
dynamics, decoder conditioning, or nonidentifiability
\citep{bowman2016generating,chen2017vlae,wang2021noniden,dang2024cvaecollapse};
a further reading attributes the neglect to a decoder that sees the ground
truth throughout training \citep{defauw2019hierarchical}, or to data that pair
each input with a single recorded future \citep{zhou2021vae2}.
Proposed mitigations include KL scheduling, free bits, auxiliary objectives,
and restrictions on decoder inputs
\citep{sonderby2016ladder,kingma2016iaf,goyal2017zforcing,roberts2018musicvae,miladinovic2022dropout}.
Closely related, \citet{sip2025doubleprojection} examine how teacher-forcing
intervals affect the use of stochastic inputs in dynamical VAEs. We study this
interaction in physical-time world models, distinguishing sampling-related
accuracy gains from trajectory-associated latent utilization through posterior
specificity and latent swapping.

\paragraph{Rollout supervision.}
The mismatch between training on observed states and predicting from generated
states is known as exposure bias \citep{ranzato2016mixer} and motivates scheduled
sampling \citep{bengio2015scheduled,huszar2015hownot}, data aggregation
\citep{venkatraman2015dad,talvitie2017selfcorrecting}, adversarial matching
\citep{lamb2016professor}, prediction-dependent transitions in recurrent world
models \citep{chiappa2017res}, and training with self-generated context
\citep{chen2024diffusionforcing,huang2025selfforcing}. Latent overshooting
extends variational training to multi-step predictions
\citep{hafner2019planet}, while RDR supervises decoded free-running
trajectories in a deterministic latent world model \citep{shah2026rdr}. Our
focus is the effect of rollout supervision on a trajectory-level variational
latent, beyond improvements in prediction accuracy. Re-anchoring experiments
vary access to ground-truth states while holding the rollout posterior fixed.
Additional implementation differences between teacher-forced and rollout
training are documented in Appendix~\ref{app:impl}.

\section{\method{}: a stochastic physical-time flow}

\subsection{Physical-time flow under teacher-forced supervision}
\label{sec:setup}

Given an initial observation $o_0$ and instruction $\ell$, a frozen image
encoder produces tokens $s_t = \mathrm{enc}(o_t)$, and a decoupling module maps
$(s_0, s_t)$ to $z_t = \mathrm{dec}(s_0, s_t)$. A frozen instruction encoder
provides $z_g$, and a frozen decoder reconstructs frames. ODEWorld's
physical-time flow (PT-Flow) evolves these latents through
\begin{equation}
\frac{\mathrm{d}z}{\mathrm{d}\tau} = v_\theta(z_0, z, z_g, \tau),
\qquad z(0) = z_0 ,
\label{eq:ode}
\end{equation}
A single numerical integration generates the trajectory, with readouts at
arbitrary query times. We use fixed-step Runge--Kutta integration at inference
(Appendix~\ref{app:impl}). Teacher-forced training evaluates the field at
demonstration states $z^{\star}_{\tau}$ and matches a target velocity $\dot
z^{\star}_{\tau}$:
\begin{equation}
\mathcal{L}_{\mathrm{tf}} \;=\;
\big\| v_\theta\big(z_0, \underbrace{z_\tau^{\star}}_{\text{ground truth}}, z_g, \tau\big)
- \dot z_\tau^{\star} \big\|^2 .
\label{eq:tf}
\end{equation}
Appendix~\ref{app:impl} details target construction and differences from the
released model. Crucially, the field receives ground-truth states during
training but its own predictions during inference.

For fixed initial conditions and goal, the deterministic ODE produces one
trajectory. Where its inputs do not distinguish alternative continuations,
squared-error supervision favors their conditional mean velocity
\citep{bishop1994mdn,mathieu2016beyondmse}, which can yield a rollout
inconsistent with either alternative.

\subsection{The trajectory-level latent and the supervision shortcut}
\label{sec:latent}

To represent multiple futures, we condition the field on $\epsilon$, held
fixed throughout each rollout: $v_\theta(z_0, z, z_g, \epsilon, \tau)$. Under
teacher forcing, the posterior $q(\epsilon \mid z_0, z^{\star}_{\tau}, z_g,
\tau)$ infers the latent from a demonstration state; a learned prior
$p(\epsilon \mid z_0, z_g)$ supplies samples at inference. A KL penalty
regularizes the posterior toward the prior.

However, the field is evaluated at $z^{\star}_{\tau}$
(Figure~\ref{fig:method}a), and that state can already identify the
demonstrated continuation, allowing the field to predict its velocity without
informative latent conditioning. This possible shortcut reduces the incentive to encode
continuation-specific information in $\epsilon$. It does not guarantee latent
neglect, as our low-dimensional example shows (Appendix~\ref{app:toy}). In our
main experiments, however, posterior means vary little across trajectories,
and swapping inferred latents provides no measurable accuracy change
(Section~\ref{sec:mech}). We call this \emph{decorative latent conditioning}:
sampling improves best-of-$K$ accuracy despite limited evidence of
trajectory-specific encoding and use.

\subsection{Rollout matching}
\label{sec:roll}

\method{} instead supervises freely integrated trajectories. For readout times
$\tau_1 < \cdots < \tau_G$ and demonstration states $z^{\star}_{1:G}$, a
permutation-invariant posterior pools the state--time pairs and produces one
sample:
\begin{equation}
q\big(\epsilon \,\big|\, z_0, z_g, \{(z^{\star}_i, \tau_i)\}_{i=1}^{G}\big),
\end{equation}
Conditioned on this sample, the field integrates from $z_0$ without receiving
intermediate ground-truth states. We match its predictions $\hat
z(\tau_i;\epsilon)$ to the demonstration:
\begin{equation}
\mathcal{L}_{\mathrm{roll}} \;=\;
\frac{1}{G}\sum_{i=1}^{G}\big\| \hat z(\tau_i;\epsilon) - z^{\star}_i \big\|^2
\;+\; \beta \, \mathrm{KL}\!\left[q \,\|\, p\right] ,
\label{eq:roll}
\end{equation}
We use $G=4$ readout times per trajectory, and the KL regularizer uses a
per-dimension free-bits floor.
Solver settings, hyperparameters, and differences in posterior inputs and KL
aggregation between training variants are documented in
Appendix~\ref{app:impl}.

Conditional on $z_0$ and $z_g$, $\epsilon$ is the only input through which the
demonstrated future enters the rollout (Figure~\ref{fig:method}b). Removing
intermediate ground-truth conditioning therefore encourages the posterior to
encode information useful for reconstructing that future. At inference, we sample $\epsilon$ from the prior, integrate once, and decode.
Training and inference share the frozen modules and velocity field in
Figure~\ref{fig:method}c; the posterior supplies the training latent and the
prior supplies the inference latent. Solver settings differ as detailed in
Appendix~\ref{app:impl}. Each sampled
trajectory uses the same number of velocity-field evaluations as PT-Flow, with
comparable measured latency (Section~\ref{sec:main}); generating $K$ samples
requires $K$ integrations.

\subsection{Posterior specificity and the latent swap}
\label{sec:diag}

Because best-of-$K$ gains alone do not establish informative latent
conditioning, we apply two complementary diagnostics to trained models.
Inspired by the active-units statistic of \citet{burda2016iwae}, posterior
specificity measures the across-trajectory standard deviation of each
posterior mean, normalized by the corresponding average prior standard
deviation and averaged over dimensions. This normalization accounts for our
learned conditional prior. Low specificity indicates limited variation in
posterior means, but does not exclude information in other posterior
statistics.

Inspired by latent control swapping \citep{seitzer2024dyst}, the latent swap
compares rollouts conditioned on latents inferred from the target trajectory
and an unrelated trajectory, holding the initial state and goal fixed. Both
are scored against the same demonstration. An advantage for the matched latent
supports the use of trajectory-associated information; equal scores indicate
no measurable benefit under this intervention, rather than independence from
$\epsilon$. Such information may include task, scene, or timing differences;
this intervention does not isolate continuation choice under matched initial
conditions and instructions. Both diagnostics require one evaluation pass
over held-out trajectories and no retraining. Section~\ref{sec:mech} reports
thresholds, sample sizes, and results.

\section{Experiments}

\subsection{Setup}
\label{sec:exp-setup}

We evaluate on LIBERO \citep{liu2023libero}, using $5{,}000$ trajectories
across $100$ tasks, re-rendered at $256\times256$ by replaying stored
simulator states. We allocate $4{,}500$ trajectories to training and $250$
each to fixed development and test sets; the test set is excluded from model
development. Unless otherwise stated, models train for $30$k steps, evaluation
uses the full test set with $K=10$ conditional-prior samples per stochastic
model, and results report the mean and standard deviation across three seeds. Seed
spreads are normalized by $n$ rather than $n-1$, since they describe the
dispersion of three runs rather than estimate a population, and the one
development-split result is labeled where it appears.
We retrain the released ODEWorld architecture using Equation~\ref{eq:tf},
since the release provides inference code only; all models share the same
frozen encoders and decoder.
Training comparisons match optimizer steps; training NFE and GPU-hours are
not reported, so these comparisons do not establish equal-compute performance.

PSNR is evaluated against recorded observations at mid-trajectory and final
frames, reporting both best-of-$K$ and mean sample performance. Best-of-$K$
follows the selection principle of variety loss \citep{gupta2018socialgan} and
may favor dispersed predictions \citep{thiede2019varietyloss}, motivating
complementary mean reporting. These frame-level metrics do not establish
whole-trajectory consistency. Diversity is the mean pairwise distance between
terminal latent states and measures representation-space variation. All
frames are compared after decoding with the shared frozen decoder, so
differences reflect the predicted latent trajectory rather than the decoder. LPIPS and FVD are reported in
Section~\ref{sec:perc}; implementation details and baseline deviations are
documented in Appendix~\ref{app:impl}.

\subsection{Accuracy and diversity at comparable inference cost}
\label{sec:main}

\begin{table}[t]
\centering
\scriptsize
\setlength{\tabcolsep}{5pt}
\caption{Results on the LIBERO test set ($250$ trajectories), with $K=10$
samples. PSNR is reported in dB; diversity denotes mean pairwise terminal
latent distance. Both training budgets use three seeds; standard deviations
are shown for $30$k steps, and means only for $60$k steps. The rollout control
is trained with $\epsilon=0$ but retains random conditioning at
evaluation, as discussed in the text. Bold indicates the
highest PSNR within each training budget.}
\label{tab:main}
\begin{tabular}{lccccc}
\toprule
& \multicolumn{2}{c}{mid-trajectory} & \multicolumn{2}{c}{final frame} & \\
\cmidrule(lr){2-3}\cmidrule(lr){4-5}
Model & best-of-10 & mean & best-of-10 & mean & diversity \\
\midrule
ODEWorld (PT-Flow)                   & $17.81${\scriptsize$\pm.22$} & $17.81${\scriptsize$\pm.22$} & $15.98${\scriptsize$\pm.17$} & $15.98${\scriptsize$\pm.17$} & $0.00$ \\
\;+ latent, teacher-forced           & $18.22${\scriptsize$\pm.15$} & $17.86${\scriptsize$\pm.23$} & $16.20${\scriptsize$\pm.24$} & $15.87${\scriptsize$\pm.26$} & $1.46$ \\
\;+ rollout control                 & $19.33${\scriptsize$\pm.06$} & $\mathbf{19.02}${\scriptsize$\pm.05$} & $18.18${\scriptsize$\pm.20$} & $\mathbf{17.77}${\scriptsize$\pm.19$} & $0.51$ \\
\textbf{\method{}} (both)            & $\mathbf{19.82}${\scriptsize$\pm.10$} & $18.85${\scriptsize$\pm.10$} & $\mathbf{18.42}${\scriptsize$\pm.08$} & $17.29${\scriptsize$\pm.10$} & $\mathbf{2.03}$ \\
\midrule
\multicolumn{6}{l}{\textit{$60$k steps, three seeds per row}} \\
ODEWorld (PT-Flow)                   & $18.67$ & $18.67$ & $16.35$ & $16.35$ & $0.00$ \\
\;+ latent, teacher-forced           & $18.74$ & $18.45$ & $16.69$ & $16.35$ & $1.22$ \\
\;+ rollout control                 & $19.61$ & $\mathbf{19.16}$ & $18.64$ & $\mathbf{18.07}$ & $0.51$ \\
\textbf{\method{}} (both)            & $\mathbf{19.80}$ & $19.07$ & $\mathbf{18.74}$ & $17.78$ & $1.18$ \\
\bottomrule
\end{tabular}
\end{table}

At $30$k steps, \method{} improves best-of-10 PSNR over ODEWorld by $2.01$ and
$2.44$\,dB at the mid-trajectory and final frames, respectively
(Table~\ref{tab:main}). Rollout matching without learned latent conditioning
recovers $1.52$ and $2.20$\,dB, while \method{} exceeds this control by
$0.49$ and $0.24$\,dB. Mean sample PSNR also exceeds ODEWorld by $1.04$ and
$1.31$\,dB, although it remains $0.17$ and $0.48$\,dB below the rollout
control. At $60$k steps, the best-of-10 advantage over ODEWorld persists at
$1.13$ and $2.39$\,dB. Learned latent conditioning increases terminal diversity to $2.03$, compared
with $1.46$ under teacher forcing and $0.51$ for the rollout control; ODEWorld
has none, since a deterministic field integrates to one trajectory per initial
state and goal. The control's residual diversity arises from evaluation
with an untrained prior despite training at $\epsilon=0$; it is therefore not
strictly deterministic. The reported increments are relative to this control;
they do not isolate the latent's contribution against rollout evaluation at
$\epsilon=0$.

\begin{figure}[t]
\centering
\includegraphics[width=\linewidth]{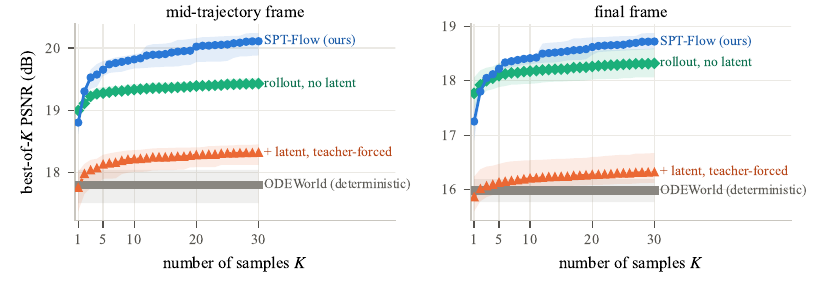}
\caption{Best-of-$K$ PSNR as a function of sample count at $30$k training
steps. Lines show means across three seeds, and shaded bands indicate the
minimum-to-maximum seed range. Rollout matching improves performance at $K=1$,
while \method{} exhibits larger gains as $K$ increases. ODEWorld is
independent of $K$.}
\label{fig:bestofk}
\end{figure}

The sampling curves further distinguish the two contributions
(Figure~\ref{fig:bestofk}): rollout matching improves accuracy at $K=1$, by $1.2$ and $1.8$\,dB at the two
frames, while increasing $K$ from $1$ to $30$ yields gains of $1.30$ and
$1.47$\,dB for \method{}, versus $0.43$ and $0.55$\,dB for the control. At the final frame,
\method{} overtakes the control at $K=3$ and achieves a $0.41$\,dB advantage at
$K=30$.

Per-sample inference costs remain comparable: a $50$-frame latent rollout
takes $151.1$\,ms for \method{} versus $149.9$\,ms for ODEWorld on the same
GPU, an approximately $0.8\%$ increase, with $200$ velocity-field evaluations
for both. The latent is sampled once and held fixed throughout integration.
Best-of-$K$ evaluation nevertheless requires $K$ rollouts, with computational
cost scaling approximately linearly under sequential execution.

\subsection{Latent specificity and utilization}
\label{sec:mech}

\begin{table}[t]
\centering
\scriptsize
\setlength{\tabcolsep}{6pt}
\caption{Both diagnostics of Section~\ref{sec:diag} are applied to trained
models on held-out trajectories. Specificity measures
posterior-mean variation normalized by the prior scale; the threshold $0.25$
was fixed before evaluation. \emph{Own} and \emph{swapped} use latents inferred
from the target trajectory and another trajectory, respectively. PSNR is
measured at the true temporal midpoint and is comparable within this
evaluation protocol. At $30$k steps, PSNR uses $250$ trajectories across three
seeds; specificity ranges use $40$ trajectories. The $60$k and low-floor
settings use seed $0$ and $40$ trajectories.}
\label{tab:mech}
\begin{tabular}{lccccc}
\toprule
Training & specificity & verdict & prior best-of-10 & own & swapped \\
\midrule
teacher-forced, $30$k & $0.007$--$0.012$ & not specific $(0/3)$ & $17.71$ & $17.34$ & $17.34$ \\
teacher-forced, $60$k & $0.010$ & not specific & $18.37$ & $18.04$ & $18.04$ \\
teacher-forced, low KL floor & $0.002$ & not specific & $17.56$ & $17.10$ & $17.10$ \\
\midrule
rollout matching, $30$k & $\mathbf{0.306}$--$\mathbf{0.376}$ & specific $(3/3)$ & $19.45$ & $\mathbf{18.96}$ & $17.91$ \\
\bottomrule
\end{tabular}
\end{table}

Teacher-forced training yields posterior specificity of $0.007$--$0.012$,
corresponding to posterior-mean variation of approximately $1\%$ of the prior
scale (Table~\ref{tab:mech}). Latents inferred from the target and unrelated
trajectories produce indistinguishable PSNR at the reported precision across
all three seeds. Neither doubling the training budget nor lowering the KL
floor changes this pattern in the tested configurations. In contrast, rollout
matching exceeds the predefined specificity threshold of $0.25$ in every seed,
with all $32$ dimensions above $0.29$, whereas no dimension of the
teacher-forced posterior exceeds $0.022$ in any seed
(Appendix~\ref{app:diag}). Using the target trajectory's inferred latent improves PSNR by $1.05$\,dB over
the swapped latent, with a per-seed range of $0.78$ to $1.19$\,dB
(Table~\ref{tab:anchor}), providing evidence that trajectory-associated
information contributes to prediction.

These results support the distinction between decorative and informative
latent conditioning introduced in Section~\ref{sec:latent}: best-of-$K$ gains
can coexist with weak posterior specificity and no measurable advantage from
the trajectory-matched latent. However, the diagnostics do not establish that
the latent selects among distinct continuations within a task. The association
between task-level branching and the latent's accuracy gain is weak, with
Spearman correlations of $0.005$ and $0.115$ at the mid-trajectory and final
frames (Appendix~\ref{app:diag}). We therefore interpret the results as
evidence of trajectory-associated latent utilization, without attributing that
information specifically to within-task continuation selection.

\subsection{Perceptual and distributional quality}
\label{sec:perc}

\begin{table}[t]
\centering
\scriptsize
\setlength{\tabcolsep}{6pt}
\caption{Perceptual and distributional evaluation across three seeds (mean
$\pm$ standard deviation; lower is better). LPIPS \citep{zhang2018lpips} is
evaluated at the mid-trajectory frame on $50$ held-out trajectories with
$K=10$. FVD \citep{unterthiner2018fvd} uses $200$ recorded reference clips and
$50$ generated clips for ODEWorld or $100$ for each other model. The rollout
control has no learned trajectory latent but retains random conditioning
during evaluation.}
\label{tab:perc}
\begin{tabular}{lccc}
\toprule
& ODEWorld & rollout control & \method{} \\
\midrule
LPIPS, mean of $K$   & $0.181${\scriptsize$\pm.008$} & $\mathbf{0.129}${\scriptsize$\pm.001$} & $0.139${\scriptsize$\pm.003$} \\
LPIPS, best of $K$   & $0.181${\scriptsize$\pm.008$} & $0.122${\scriptsize$\pm.001$} & $\mathbf{0.116}${\scriptsize$\pm.004$} \\
FVD                  & $265${\scriptsize$\pm41$} & $210${\scriptsize$\pm5$} & $\mathbf{187}${\scriptsize$\pm9$} \\
\bottomrule
\end{tabular}
\end{table}

Perceptual evaluation shows a pattern consistent with the PSNR results
(Table~\ref{tab:perc}). The rollout control reduces mean LPIPS from $0.181$ to
$0.129$, a $29\%$ improvement over ODEWorld. Adding learned latent conditioning further reduces best-of-$K$ LPIPS from
$0.122$ to $0.116$, a $36\%$ reduction against ODEWorld, while increasing mean
LPIPS by $0.010$. With both rollout-based models evaluated
using $100$ generated clips, \method{} has a lower estimated FVD, $187$ versus $210$,
approximately $11\%$. These results indicate that rollout matching improves
average perceptual accuracy, while learned latent conditioning improves
best-of-$K$ performance. The FVD estimates are exploratory distributional
measurements and do not establish semantic mode coverage.
Comparisons with ODEWorld's FVD remain qualified by the unequal generated
sample counts, $50$ clips against $100$ for each rollout-trained model,
against the same $200$ recorded references.

\subsection{Training budget and supervision ablations}
\label{sec:ablation}

\begin{table}[t]
\centering
\scriptsize
\setlength{\tabcolsep}{3.5pt}
\caption{Ground-truth states are withdrawn from the training rollout in three
stages. Rollout variants share the same posterior; teacher forcing follows
Equation~\ref{eq:tf}. Specificity is reported as the mean [minimum, maximum]
across three seeds. PSNR and diversity are seed means. The swap gap is
own-minus-swapped mid-trajectory PSNR in dB, measured on seed $0$ except for
\method{}, which reports the three-seed mean and range.}
\label{tab:anchor}
\begin{tabular}{lccccc}
\toprule
Ground truth enters at & specificity & swap gap & mid best-of-10 & final best-of-10 & div. \\
\midrule
every training signal (teacher forced) & $0.009$ {\tiny$[.007,.012]$} & $0.00$ & $18.22$ & $16.20$ & $1.46$ \\
every readout of the rollout           & $0.236$ {\tiny$[.228,.245]$} & $1.07$ & $19.94$ & $18.58$ & $2.67$ \\
every other readout                    & $0.298$ {\tiny$[.243,.359]$} & $0.97$ & $19.72$ & $18.55$ & $2.95$ \\
never (\method{})                      & $0.332$ {\tiny$[.306,.376]$} & $1.05$ {\tiny$[.78,1.19]$} & $19.82$ & $18.42$ & $2.03$ \\
\bottomrule
\end{tabular}
\end{table}

\paragraph{Training budget.}
At $60$k steps, teacher-forced latent conditioning improves mid-trajectory
best-of-10 PSNR over ODEWorld by only $0.07$\,dB, whereas \method{} retains a
$1.13$\,dB advantage (Table~\ref{tab:main}). Its diversity decreases from
$2.03$ to $1.18$, and the rollout control accounts for a larger share of the
improvement: $0.94$ of $1.13$\,dB, compared with $1.52$ of $2.01$\,dB at $30$k
steps. At $120$k steps, the mid-trajectory advantage over ODEWorld remains $0.97$\,dB
in one seed and $1.53$\,dB in the other, with final-frame advantages of $2.07$
and $2.37$\,dB (Appendix~\ref{app:long}). A similar decline occurs
on the ten-task subset, with gains of $1.36$, $1.15$ and $0.84$\,dB at $10$k,
$20$k and $30$k steps (Appendix~\ref{app:tentask}). This subset entails
greater data reuse but also changes task coverage, so it does not isolate the
effect of repeated exposure.

\paragraph{Ground-truth re-anchoring.}
Reducing access to intermediate ground-truth states increases mean posterior
specificity from $0.236$ to $0.298$ and $0.332$ across the rollout variants
(Table~\ref{tab:anchor}). Only fully free-running training exceeds the $0.25$
threshold in all three seeds. Nevertheless, re-anchoring at every readout
already yields a $1.07$\,dB swap gap, indicating latent utilization even below
this threshold. The swap gap does not increase monotonically with specificity,
and the most frequently anchored rollout achieves the highest PSNR. These
results distinguish posterior variation, latent utilization, and prediction
accuracy. Because the rollout variants share their posterior configuration,
their specificity trend supports the proposed supervision-shortcut
interpretation; comparisons with teacher forcing additionally involve the
implementation differences documented in Appendix~\ref{app:impl}.

\begin{table}[t]
\centering
\scriptsize
\setlength{\tabcolsep}{5pt}
\caption{Each variant changes one setting relative to the reference: $\beta = 3\times10^{-3}$, latent dimension $32$, and
$8$ Euler steps during training. Variants use one seed; the reference reports
three-seed means and the specificity range. PSNR is in dB.}
\label{tab:hp}
\begin{tabular}{lccccc}
\toprule
Configuration & mid best-of-10 & mid mean & final best-of-10 & div. & specificity \\
\midrule
$\beta = 10^{-3}$        & $19.79$ & $18.83$ & $18.36$ & $1.56$ & $0.44$ \\
reference ($\beta = 3\times10^{-3}$) & $19.82$ & $18.85$ & $18.42$ & $2.03$ & $0.31$--$0.38$ \\
$\beta = 10^{-2}$        & $19.53$ & $18.82$ & $17.98$ & $1.06$ & $0.20$ \\
latent dimension $16$    & $19.81$ & $18.71$ & $\mathbf{18.64}$ & $2.66$ & $\mathbf{0.58}$ \\
$16$ Euler steps         & $\mathbf{19.85}$ & $\mathbf{19.13}$ & $18.34$ & $0.95$ & $0.26$ \\
\bottomrule
\end{tabular}
\end{table}

\begin{table}[t]
\centering
\scriptsize
\setlength{\tabcolsep}{6pt}
\caption{Temperature $T$ scales the prior standard deviation at inference,
without retraining; $T=1$ is the default and \method{} is trained for $30$k
steps. PSNR is
reported as mean $\pm$ standard deviation across three seeds, and diversity as
the seed mean.}
\label{tab:temp}
\begin{tabular}{lccccc}
\toprule
$T$ & mid best-of-10 & mid mean & final best-of-10 & final mean & div. \\
\midrule
$0.5$ & $19.53${\scriptsize$\pm.07$} & $\mathbf{18.95}${\scriptsize$\pm.09$} & $18.12${\scriptsize$\pm.09$} & $\mathbf{17.40}${\scriptsize$\pm.09$} & $1.08$ \\
$0.7$ & $19.68${\scriptsize$\pm.09$} & $18.91${\scriptsize$\pm.09$} & $18.25${\scriptsize$\pm.08$} & $17.36${\scriptsize$\pm.10$} & $1.48$ \\
$1.0$ & $19.82${\scriptsize$\pm.10$} & $18.85${\scriptsize$\pm.10$} & $18.42${\scriptsize$\pm.08$} & $17.29${\scriptsize$\pm.10$} & $2.03$ \\
$1.3$ & $\mathbf{19.91}${\scriptsize$\pm.10$} & $18.77${\scriptsize$\pm.10$} & $\mathbf{18.51}${\scriptsize$\pm.08$} & $17.21${\scriptsize$\pm.11$} & $\mathbf{2.52}$ \\
\bottomrule
\end{tabular}
\end{table}

\paragraph{Hyperparameter sensitivity.}
Mid-trajectory best-of-10 PSNR ranges from $19.53$ to $19.85$\,dB across the
tested configurations, while posterior specificity varies more substantially
(Table~\ref{tab:hp}). Increasing $\beta$ from $10^{-3}$ to $10^{-2}$ reduces
specificity from $0.44$ to $0.20$, with the swap gap decreasing to
$0.38$\,dB at the higher weight. A $16$-dimensional latent achieves the
highest final-frame best-of-10 PSNR, whereas $16$ Euler steps yield the
highest mid-trajectory mean PSNR and lowest diversity. These comparisons are
exploratory because each non-reference setting uses one seed. Reducing the
effective batch size from $16$ to $8$ also produces near-zero specificity
($0.004$--$0.005$) and zero diversity in both tested seeds, indicating
sensitivity to the optimization configuration.

\paragraph{Sampling temperature.}
Scaling the prior standard deviation trades mean accuracy against best-of-10
accuracy and diversity without retraining (Table~\ref{tab:temp}). Lower temperatures improve mean PSNR but reduce both best-of-10 PSNR and
diversity, which falls from $2.52$ at $T=1.3$ to $1.08$ at $T=0.5$.
At $T=0.5$, final-frame best-of-10 PSNR falls slightly below the rollout
control ($18.12$ versus $18.18$), whereas $T=0.7$ already exceeds it
($18.25$). None of the tested temperatures matches the control's final-frame
mean PSNR of $17.77$. Relative to $T=1$, increasing the temperature to $1.3$
improves best-of-10 PSNR by $0.09$\,dB at both frames while reducing mean PSNR
by $0.08$\,dB. The default therefore represents an intermediate setting within
the evaluated range.

\paragraph{KL-based mitigations.}
The tested KL adjustments do not recover trajectory-associated latent
utilization under teacher forcing (Appendix~\ref{app:mitig}). KL annealing, a
threefold reduction in KL weight, and a fourfold increase in the free-bits
floor yield specificity values of $0.010$--$0.027$ and swap gaps no larger
than $0.001$\,dB. These results indicate that the failure is not resolved by
the tested regularization settings and are consistent with the
supervision-shortcut hypothesis.

\section{Conclusion}

We introduced \method{}, a stochastic extension of ODEWorld that combines a
trajectory-level latent variable with rollout matching while retaining a
single physical-time integration per sampled trajectory. On LIBERO, at $30$k steps, it
improves best-of-10 PSNR over our retrained ODEWorld baseline by $2.0$ and
$2.4$\,dB at the mid-trajectory and final frames, respectively, and reduces
mean LPIPS by $23\%$, with comparable per-sample inference latency. The
prediction advantage persists when both models receive a fourfold increase in
training steps. The rollout control recovers most of the accuracy improvement;
learned latent conditioning increases diversity and best-of-$K$ performance,
with lower mean sample accuracy relative to that control.

Our results also distinguish predictive accuracy from latent utilization.
Under teacher-forced supervision, best-of-$K$ accuracy can improve despite low
posterior specificity and no measurable PSNR benefit from trajectory-matched latents.
Rollout matching exceeds the predefined specificity threshold across all three
seeds, and the corresponding inferred latent outperforms a swapped latent by
$1.05$\,dB. Re-anchoring experiments further associate reduced access to
intermediate ground-truth states with higher posterior specificity.

\paragraph{Scope and limitations.}
These findings motivate reporting latent diagnostics alongside prediction
metrics. They support trajectory-associated information use, while leaving
same-condition semantic mode coverage unresolved. Evaluation covers held-out
trajectories from tasks represented in training; generalization to unseen
tasks and downstream control benefits remain untested. Latent utilization
also depends on the optimization setting, including effective batch size
(Section~\ref{sec:ablation}). The training comparisons match steps, and the
rollout control retains random evaluation inputs, limiting claims about
compute efficiency and the isolated contribution of learned conditioning.

\clearpage
\section*{Reproducibility Statement}

Section~\ref{sec:exp-setup} describes the data splits, evaluation protocols,
and reporting across random seeds. Appendix~\ref{app:impl} documents data
preparation, model architectures, hyperparameters, training procedures,
numerical solvers, and implementation differences between comparison models.
Diagnostic definitions and thresholds appear in Sections~\ref{sec:diag}
and~\ref{sec:mech}, with additional analyses in Appendix~\ref{app:diag}.
Hyperparameter and sampling-temperature sensitivity analyses are reported in
Section~\ref{sec:ablation} and Appendix~\ref{app:hp}; extended-budget results
appear in Appendix~\ref{app:long}. Development-set results are explicitly
identified in Appendix~\ref{app:tentask}. Training and evaluation code is
included in the supplementary material and will be publicly released.

\section*{AI Use Statement}

Generative AI tools assisted with language editing and with reviewing the
clarity and scope of manuscript claims. The authors reviewed all AI-assisted
revisions and take full responsibility for the final manuscript, including its
claims, citations, and experimental reporting.

\bibliographystyle{iclr2027_conference}
\bibliography{refs}

\appendix

\section{Implementation details}
\label{app:impl}

\paragraph{Encoders and latent.} The image encoder is a frozen
DINOv2-with-registers backbone at $224\times224$; the language encoder is
frozen. The decoupling module is a four-layer cross-attention block producing a
single $768$-dimensional latent token. The velocity field is a three-layer MLP
of width $4096$ with FiLM conditioning on a learned sinusoidal embedding of
$\tau$. The posterior and prior are two-layer MLPs of width $1024$ and $512$
producing a $32$-dimensional Gaussian.

\paragraph{Conditioning.} We condition on the instruction rather than on a
goal image. Supplying the final frame determines most of the continuation from
the conditioning alone, which leaves little for a trajectory-level latent to
represent; the instruction leaves the choice of continuation open, which is
the setting the diagnostics of Section~\ref{sec:diag} are about.

\paragraph{Optimization.} AdamW, learning rate $10^{-4}$ with $500$ warm-up
steps and cosine decay to zero over the declared step budget, weight decay
$10^{-4}$, gradient clipping at $1.0$, batch of four trajectories with four
sampled times each, distributed over four A100 or A800 GPUs (effective
batch $16$; Section~\ref{sec:ablation} shows why this matters). The first $1000$
steps train the reconstruction term only. Rollout matching integrates with
eight Euler steps between consecutive sampled times; inference uses a
fixed-step fourth-order Runge--Kutta solver with $200$ evaluations for a
$50$-frame trajectory.

Because the learning rate anneals over the declared budget, a model taken partway through a $60$k-step run is not comparable to the final
model of a $30$k-step run. Every comparison in the main text is between runs at their own final step; the sweep of Appendix~\ref{app:tentask}
compares the same runs at matched intermediate steps, which is
a comparison across arms at each step and not across steps.

\paragraph{Differences between the teacher-forced and rollout arms.} The two
arms share the encoder, decoder, latent dimension, KL weight and free-bits
floor, but they are not identical in every other respect, and we list the
differences here so that the comparison can be read for what it is. In the
teacher-forced arm the posterior reads $(z_0, z^{\star}_\tau, z_g, \tau)$ at
the single sampled time and a fresh $\epsilon$ is drawn for each sampled time;
in the rollout arm it reads all $G=4$ readout latents through a
permutation-invariant pooling and one $\epsilon$ is shared by the four
readouts. The free-bits floor is applied per dimension in both arms, to the
batch-averaged KL in the teacher-forced arm and per trajectory before
averaging in the rollout arm. We did not run a teacher-forced arm with the
pooled posterior, so the main-text attribution to the supervision rests on
the re-anchoring series of Table~\ref{tab:anchor}, which keeps the rollout
posterior fixed. Our retraining of the base objective also lets gradients
flow through the Jacobian-vector-product target, whereas
\citet{liu2026odeworld} describe a stop-gradient; the released code ships
inference only, so we could not run the original training objective as is.

\paragraph{KL term.} $\beta = 3\times10^{-3}$ with a per-dimension floor of
$0.05$ applied before summation. The floor is required
(Section~\ref{sec:ablation}); the value was selected on the development half.

\paragraph{Dataset.} LIBERO demonstrations are re-rendered at $256\times256$ by
restoring each stored simulator state and rendering offscreen, rather than by
replaying actions, so the rendered trajectory matches the recorded one exactly.
This yields $5{,}000$ trajectories over $100$ tasks. The held-out set contains
every tenth demonstration of each task.

\paragraph{Evaluation.} PSNR, LPIPS \citep{zhang2018lpips} and SSIM \citep{wang2004ssim} are computed on decoded frames at
a mid-trajectory index and at the final index. Best-of-$K$ and the mean over
$K$ samples follow the protocol of stochastic video prediction
\citep{babaeizadeh2018sv2p,denton2018svg}, and diversity is the mean pairwise
distance between samples as in \citet{lee2018savp}, taken here between
terminal latents. FVD \citep{unterthiner2018fvd} uses a standard I3D feature extractor on
$16$-frame clips and is biased toward per-frame quality over temporal realism
\citep{ge2024fvdbias}, so we read it as one number of three rather than a
summary. SSIM moves with LPIPS and is omitted from
Table~\ref{tab:perc} for space: over the same three seed pairings, ODEWorld
$0.743$, rollout training without the latent $0.789$ and \method{} $0.792$
(best of $K$) at the mid-trajectory frame, and $0.714$, $0.774$ and $0.781$
at the final frame. Perceptual and distributional metrics use at
least $50$ trajectories: on ten trajectories we measured differences that
disappeared at fifty, in both directions, so smaller samples are not reported.

\paragraph{Per-step baseline.} An untuned per-step flow-matching world model
\citep{lipman2023flowmatching,porcher2026flowwm} trained on the same encoder
and data reaches $13.5$\,dB at the mid-trajectory frame with $500$ network
evaluations, and doubling its solver steps changes this by only $0.13$\,dB;
further tuning might narrow the gap, and we do not report it in the main
tables.

\section{Cross-scale check on ten tasks}
\label{app:tentask}

\begin{table}[h]
\centering
\scriptsize
\setlength{\tabcolsep}{6pt}
\caption{A $10$-task subset ($450$ training trajectories): the same runs at matched intermediate steps, one seed, mid-trajectory
best-of-10 on that subset's $40$-trajectory development split (not the test
split used elsewhere).}
\label{tab:tentask}
\begin{tabular}{lccc}
\toprule
$10$ tasks & $10$k & $20$k & $30$k \\
\midrule
ODEWorld         & $18.37$ & $18.85$ & $18.98$ \\
\;+ latent, tf   & $18.44$ & $19.28$ & $19.45$ \\
\textbf{\method{}} & $\mathbf{19.73}$ & $\mathbf{20.00}$ & $\mathbf{19.82}$ \\
\bottomrule
\end{tabular}
\end{table}

On a $10$-task subset (Table~\ref{tab:tentask}) \method{} leads at every step,
by $1.36$, $1.15$
and $0.84$\,dB. The margin shrinks as training continues, in the same direction
as the budget result. These two settings differ by an order of magnitude in
passes over the data at equal step counts, which is consistent with the erosion
being governed by data repetition rather than by step count, though we have not
isolated the two.

\section{Hyperparameter sensitivity of the fixed method}
\label{app:hp}

Table~\ref{tab:hp} in the main text changes one knob at a time from the
reported configuration, and Section~\ref{sec:ablation} discusses it. The swap
gap of the $\beta = 10^{-2}$ model is $0.38$\,dB against $1.05$ for the
reference.

The sampling-temperature sweep is Table~\ref{tab:temp} in Section~\ref{sec:ablation}.
\section{Four times the training steps}
\label{app:long}

\begin{table}[h]
\centering
\scriptsize
\setlength{\tabcolsep}{5pt}
\caption{Seed 0 of each arm trained for $120$k steps on a different accelerator
(RTX~5090; other tables use A100s, and the reference model reproduces its Table~\ref{tab:main} numbers on this node
to four decimals), $250$-trajectory
test split, $K=10$. A second seed is reported in the text. Models taken partway through share the $120$k cosine schedule and are not comparable with the
endpoints of Table~\ref{tab:main}.}
\label{tab:long}
\begin{tabular}{lcccccc}
\toprule
& \multicolumn{2}{c}{ODEWorld} & \multicolumn{4}{c}{\method{}} \\
\cmidrule(lr){2-3}\cmidrule(lr){4-7}
Step & mid & final & mid best-of-10 & mid mean & final best-of-10 & diversity \\
\midrule
$30$k  & $18.10$ & $15.87$ & $19.45$ & $18.49$ & $18.19$ & $1.56$ \\
$60$k  & $18.10$ & $15.71$ & $19.88$ & $19.17$ & $18.67$ & $0.98$ \\
$90$k  & $19.00$ & $16.81$ & $19.85$ & $19.17$ & $18.85$ & $0.83$ \\
$120$k & $19.07$ & $16.88$ & $\mathbf{20.04}$ & $\mathbf{19.37}$ & $\mathbf{18.95}$ & $0.68$ \\
\bottomrule
\end{tabular}
\end{table}

At the $120$k endpoint (Table~\ref{tab:long}) the gap is $0.97$\,dB at the
mid-trajectory frame and
$2.07$\,dB at the final frame, and the deterministic model's own curve has
flattened ($19.00$ at $90$k, $19.07$ at $120$k). The latent stays
trajectory-specific at this budget (specificity $0.31$, all $32$ dimensions
above the threshold; swap gap $0.74$\,dB against $1.05$ at $30$k), and the
best-of-$K$ slope from $K=1$ to $30$ is $0.75$\,dB against $1.30$ at $30$k,
so the longer run retains latent utilization while showing a smaller gain
from sampling. Its diversity falls from $1.56$ to $0.68$ over the run, continuing
the $2.03$ to $1.18$ drop between the $30$k and $60$k rows of
Table~\ref{tab:main}. This comparison does not include a $120$k-step rollout
control and therefore does not isolate the learned latent's incremental benefit.

A second seed reproduces the endpoint on Ampere hardware: trained on an A100
(\method{}) and an A800 (ODEWorld), with the reference model reproducing to four decimals on both nodes, the two models reach $20.47$ and $18.94$\,dB
at the mid-trajectory frame and $19.37$ and $17.00$\,dB at the final frame, a
gap of $1.53$ and $2.37$\,dB against $0.97$ and $2.07$ for seed~0. The latent
remains trajectory-specific at this budget in both seeds (specificity $0.365$
and $0.307$, all $32$ dimensions above the threshold).

\section{Mode averaging in a controlled setting}
\label{app:toy}

A field trained with first-order supervision on two-mode data integrates to a
path between the modes. Trajectories in this task start at a common point and
reach a common endpoint along an upper or a lower arc with equal probability.
The deterministic field produces neither arc. Adding unstructured Gaussian
noise to the field broadens the paths without recovering the two modes in
this experiment. A latent-conditioned field recovers both arcs. Wasserstein-1 distance
\citep{villani2009ot} between
predicted and demonstrated mid-trajectory positions, over three seeds, is
$0.350$ for the deterministic field --- exactly the arc amplitude, meaning it
predicts the midline --- $0.143$ with additive Gaussian noise, and $0.031$ and
$0.040$ for a latent-conditioned field trained with teacher forcing and with
rollout matching respectively. Both latent variants recover the modes here: in
this low-dimensional setting, where the posterior reads the whole trajectory,
even the teacher-forced latent gets used. Latent neglect is therefore not
inevitable under teacher forcing. This positive control verifies mode recovery
where modes are known, but does not isolate the effects of scale, posterior
inputs, and optimization that differ from the LIBERO setting.

\begin{figure}[h]
\centering
\includegraphics[width=\linewidth]{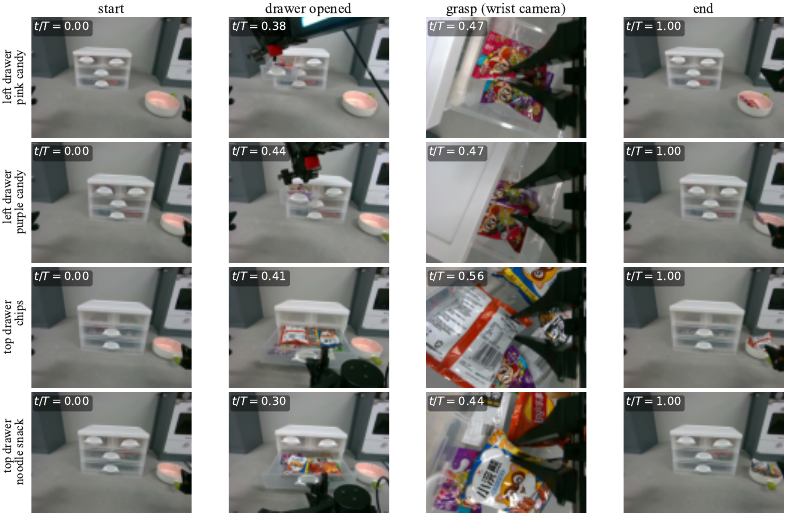}
\caption{Four teleoperated demonstrations of one instruction (\emph{open a drawer and
put a snack in the bowl}) on a real arm, one per row; fixed camera except the
wrist-camera grasp column. All start with both drawers closed and diverge in
which drawer is opened and which snack is taken. $t/T$ is normalized time.
Recorded in separate sessions; an illustration of the premise, not an
evaluation.}
\label{fig:real}
\end{figure}

The diagnostics of Section~\ref{sec:mech} read as expected where the latent
is known to be used. On $400$ fresh trajectories per seed, one of the four
latent dimensions carries the mode in every model, with a per-dimension
specificity of $0.89$--$0.98$ against the learned conditional prior (mean over
dimensions $0.31 \pm 0.02$ teacher-forced and $0.34 \pm 0.04$
rollout-trained; the mean is diluted by unused dimensions, so the
per-dimension value is the one to compare). The latent swap separates
cleanly: the mid-trajectory error of a free rollout is $0.016$ and $0.010$
with the trajectory's own latent, $0.38$ and $0.37$ with the latent of a
random other trajectory, which lies in the opposite mode about half the time
($0.70$, the full arc separation, when it does), and $0.37$ and $0.33$ with a
prior draw. Both diagnostics detect mode-associated information in this
positive control. Their low readings for the teacher-forced LIBERO model
remain specific to that model and evaluation setting. At scale the per-dimension reading
agrees with the mean: no dimension of the teacher-forced posterior exceeds
$0.022$ in any seed, and under rollout training all $32$ dimensions exceed
$0.25$ in every seed.

\section{Classical mitigations in detail}
\label{app:mitig}

Annealing the KL weight from zero over the first $5$k steps leaves the
teacher-forced posterior at specificity $0.014$ and $0.012$ across two seeds,
against $0.007$--$0.012$ without annealing and $0.31$--$0.38$ for rollout training. The
free-bits floor is load-bearing in the other direction: lowering it from
$0.05$ to $0.002$ or to zero collapses the KL to $0.014$ and $10^{-4}$, and
the low-floor row of Table~\ref{tab:mech} shows the resulting specificity.
The tested KL schedules and floors do not recover trajectory-associated
latent utilization in these teacher-forced configurations.

The additional KL settings in Table~\ref{tab:tftune} yield the same qualitative
pattern. Lowering the KL weight by a factor of three or
raising the free-bits floor fourfold moves mid-trajectory accuracy by at most
$0.20$\,dB from the reference configuration and leaves specificity between
$0.010$ and $0.027$, with the latent of a trajectory's own future scoring
within $0.001$\,dB of a shuffled one in all three configurations. These results
are consistent with the supervision-shortcut hypothesis, but do not isolate
it from the other implementation differences in Appendix~\ref{app:impl}.

\begin{table}[h]
\centering
\scriptsize
\setlength{\tabcolsep}{5pt}
\caption{Teacher-forced training with one KL knob changed; two GPUs with twice the
per-GPU batch (effective batch $16$) on the node of Appendix~\ref{app:long},
$30$k steps, seed 0, $250$-trajectory test split, $K=10$. The reference row is
retrained on the same node, so rows are comparable with each other but not
with Table~\ref{tab:main}.}
\label{tab:tftune}
\begin{tabular}{lccccc}
\toprule
Teacher-forced configuration & mid best-of-10 & mid mean & specificity & max.\ dim. & swap gap \\
\midrule
reference ($\beta=3\times10^{-3}$, floor $0.05$) & $18.52$ & $18.19$ & $0.016$ & $0.026$ & $0.001$ \\
$\beta = 10^{-3}$                                & $18.67$ & $18.34$ & $0.010$ & $0.015$ & $0.000$ \\
free-bits floor $0.2$                            & $18.72$ & $18.34$ & $0.027$ & $0.045$ & $0.001$ \\
\bottomrule
\end{tabular}
\end{table}

\section{Diagnostics}
\label{app:diag}

\paragraph{The rollout control's residual noise.}
The control is trained at $\epsilon=0$ but evaluated by the same code as the
full model, which draws $\epsilon$ from a prior the control never trained.
Residual sensitivity to this input produces its diversity of $0.51$ in
Table~\ref{tab:main} and allows best-of-$K$ gains. This control does not measure
strictly deterministic rollout performance, and its sample mean cannot be
substituted for an evaluation at $\epsilon=0$. The $0.49$\,dB mid-trajectory
increment in Section~\ref{sec:main} compares best-of-$10$ against best-of-$10$
under the reported protocol. Per seed, a paired bootstrap
\citep{efron1993bootstrap} over the $250$ test trajectories gives increments
of $+0.25$, $+0.57$ and $+0.63$\,dB, each with a $95\%$ interval excluding
zero; at the final frame one seed's interval includes zero ($+0.05$,
$[-0.10, +0.21]$). These intervals describe trajectory resampling within
each fitted seed; they do not establish robustness across unseen tasks or
additional training seeds.

\paragraph{Detail behind Table~\ref{tab:mech}.}
The per-dimension results agree with the aggregate diagnostic: teacher-forced
posterior means vary little, whereas under rollout training every one of the
$32$ dimensions exceeds $0.29$ in every seed. Thus, the latter result does
not rest on a few active dimensions. On the ten-task subset of
Appendix~\ref{app:tentask}, the rollout posterior reaches specificity $0.47$;
this result alone does not identify the information represented by the latent.

\paragraph{Exploratory association between task dispersion and prediction gain.}
We measured per-task
branching as the mean pairwise distance between demonstrations' proprioceptive
states at mid-trajectory, resampled to a common time grid, over $12$
demonstrations for each of the $100$ tasks, and correlated it with the
full model's gain over the reported rollout control per test
trajectory. Over the $250$-trajectory test split the Spearman correlation is
$0.005$ ($p=0.94$) at the mid-trajectory frame and $0.115$ ($p=0.07$) at the
final frame. On a $60$-trajectory subset the same statistic read $0.22$, which
is why we report the larger sample. These pooled estimates are exploratory:
trajectories share task-level dispersion values, and the reported analysis
does not establish task-level inferential uncertainty. The small correlations
neither establish zero association nor distinguish semantic continuation
selection from other trajectory-associated variation. More generally,
prediction quality and downstream utility require separate evaluation
\citep{zhang2026worldinworld}.

Posterior specificity is
$\operatorname{std}_{n}[\mu_q(x_n)] / \operatorname{mean}[\sigma_p]$, where
$\mu_q$ is the posterior mean for held-out trajectory $n$ and $\sigma_p$ the
prior standard deviation, averaged over latent dimensions. Both diagnostics
also record whether two distinct latents change the rollout at all, which
distinguishes an uninformative posterior from a latent that is not wired into
the field.

\begin{figure}[h]
\centering
\includegraphics[width=0.92\linewidth]{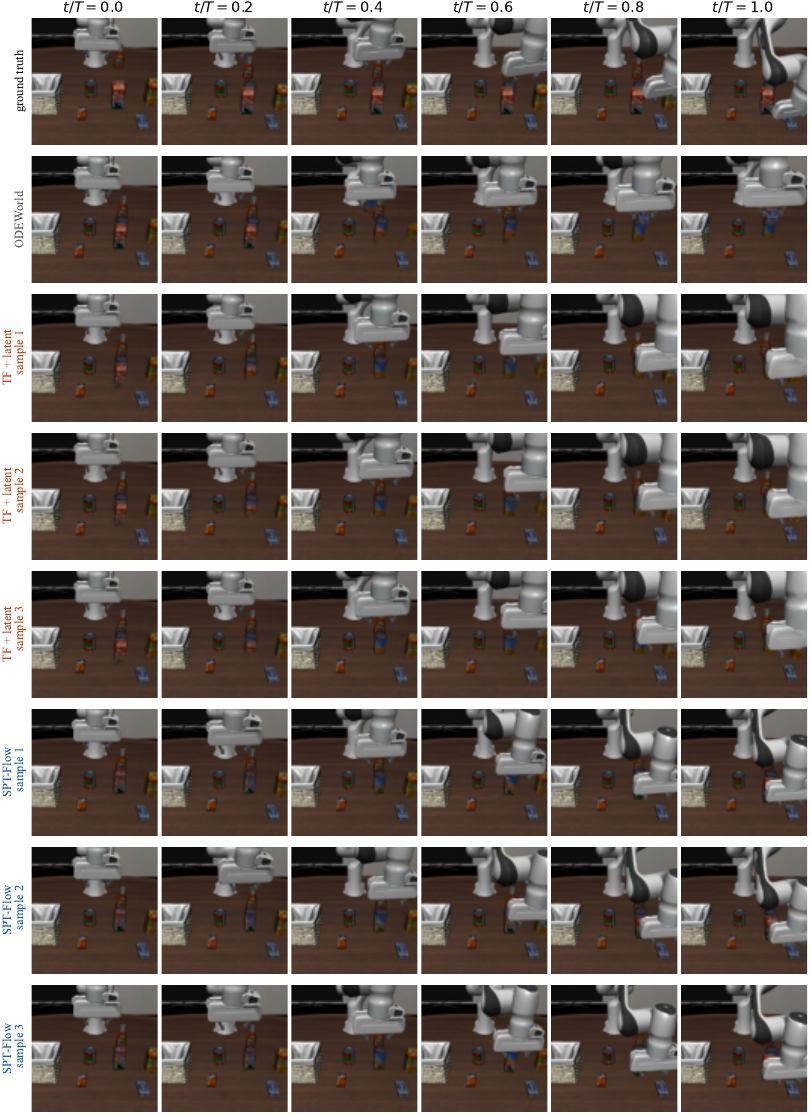}
\caption{One test trajectory (\emph{put both the cream cheese box and the
butter in the basket}), six of $50$ frames, decoded through the shared frozen
decoder. Rows: recorded continuation, ODEWorld, three samples from the
teacher-forced latent model, three samples from \method{}; all seed-$0$ models at $30$k steps from Table~\ref{tab:main}.}
\label{fig:qual}
\end{figure}

\paragraph{Qualitative rollouts.}
Figure~\ref{fig:qual} shows one test trajectory decoded from every arm of
Table~\ref{tab:main}. The three teacher-forced samples are visually identical,
which is the decorative latent of Section~\ref{sec:mech} seen directly; the
three \method{} samples differ from one another in the arm's path and follow
the recorded continuation more closely than the deterministic model. The
trajectory was chosen as one of the highest-diversity test trajectories for
\method{}, so the spread shown is toward the upper end of what the model
produces. Animated versions of four such trajectories are in the
supplementary material.

\paragraph{The recorded continuations separate in the latent once the branch is
taken.} We asked whether the frozen encoder's latent separates the four
continuations of Figure~\ref{fig:real} over time, encoding $60$
time-normalized frames per episode from the fixed camera and subtracting each
episode's first-frame latent. Within one recording session, a
leave-one-episode-out nearest-centroid classifier cannot tell the two drawers
apart reliably from the first frame ($0.54$, near chance) and then identifies the drawer
with accuracy $0.92$ from $t/T=0.05$ and $1.00$ from $t/T=0.15$. This classifier
therefore distinguishes the recorded drawer choices more accurately after
branching; its first-frame result does not establish that the representation
contains no predictive information. Each continuation was recorded as a block:
the same classifier also identifies the snack within a drawer
($0.71$--$1.00$) and the session within a continuation ($0.61$--$1.00$), and
the centroid distance between two sessions of the same continuation
($2.2$--$3.7$) matches the distance between drawers within one session
($3.2$), so session and operator style travel with the choice. Interleaving
the orders within one session would separate them.

\paragraph{Time resolution.} In the earlier $128$-pixel configuration we read
the trajectory out at $0.5\times$, $1\times$, $2\times$ and $4\times$ the
training resolution; PSNR was $18.1$, $18.05$, $18.05$ and $18.05$\,dB at the
mid-trajectory frame and $16.77$ at the final frame in all four cases. This
checks prediction consistency at the evaluated frames across readout
resolutions; it does not measure the cost of decoding additional frames.

\end{document}